**Article title**

EcoCropsAID: Economic Crops Aerial Image Dataset for Land Use Classification

**Authors**

Sangdaow Noppitak[a], Emmanuel Okafor[b], Olarik Surinta[c,*]

**Affiliations**

[a] Faculty of Science, Buriram Rajabhat University, 31000 Buri Ram, Thailand

[b] SDAIA-KFUPM Joint Research Center for Artificial Intelligence, King Fahd University of Petroleum and Minerals, Dharan 31261, Saudi Arabia

[c] Multi-agent Intelligent Simulation Laboratory (MISL) Research Unit, Department of Information Technology, Faculty of Informatics, Mahasarakham University, Mahasarakham, 44150 Thailand

**Corresponding author's email address**

Email address: olarik.s@msu.ac.th

**Abstract**

The EcoCropsAID dataset is a comprehensive collection of 5,400 aerial images captured between 2014 and 2018 using the Google Earth application. This dataset focuses on five key economic crops in Thailand: rice, sugarcane, cassava, rubber, and longan. The images were collected at various crop growth stages—early cultivation, growth, and harvest—resulting in significant variability within each category and similarities across different categories. These variations, coupled with differences in resolution, color, and contrast introduced by multiple remote imaging sensors, present substantial challenges for land use classification. The dataset is an interdisciplinary resource that spans multiple research domains, including remote sensing, geoinformatics, artificial intelligence, and computer vision. The unique features of the EcoCropsAID dataset offer opportunities for researchers to explore novel approaches, such as extracting spatial and temporal features, developing deep learning architectures, and implementing transformer-based models. The EcoCropsAID dataset provides a valuable platform for advancing research in land use classification, with implications for optimizing agricultural practices and enhancing sustainable development. This study explicitly investigates the use of deep learning algorithms to classify economic crop areas in northeastern Thailand, utilizing satellite imagery to address the challenges posed by diverse patterns and similarities across categories.

# SPECIFICATIONS TABLE

| Subject | Computer Science |
|---|---|
| **Specific subject area** | Land use refers to the classification of land based on its intended purposes, such as agriculture and water management, using satellite-captured aerial images. The cultivation of economic crops can be analyzed, planned, and managed by applying artificial intelligence algorithms, including classification and segmentation, directly to these aerial images. This dataset is also relevant to various disciplines, including deep learning, computer vision, remote sensing, data science, and computer science applications. |
| **Type of data** | Image (JPG format) |
| **Data collection** | The Thailand Economic Crops Aerial Image Dataset (EcoCropsAID) is a novel dataset featuring aerial images of the five most economically significant crops in Thailand: rice, sugarcane, cassava, rubber, and longan. The dataset was collected using the Google Earth application, selecting satellite data from 2014 to 2018 across different cultivation areas in various provinces and regions. Data collection was conducted based on information from the Agri-Map Online website, provided by the Ministry of Agriculture and Cooperatives, to ensure data accuracy. |
| **Data source location** | Country: Thailand<br>Location: northeastern region<br>Latitude: +14° 14' to +18° 27'<br>Longitude: +101° 15' to +105° 35' |
| **Data accessibility** | Repository name: Mendeley Data<br>Data identification number: 10.17632/g8fhf7fbds.2<br>Direct URL to data: https://data.mendeley.com/datasets/g8fhf7fbds/2 |
| **Related research article** | S. Noppitak and O. Surinta, Ensemble convolutional neural network architectures for land use classification in economic crops aerial images, ICIC Express Letters 15 (2021) 531–543. https://doi.org/10.24507/icicel.15.06.531 |

# VALUE OF THE DATA

- The research team used the Google Earth application to select satellite data from 2014 to 2018. Hence, the data is dynamic due to varying cultivation periods. Additionally, aerial images of economic crops were captured across different phases: cultivation, growth, and harvest. The challenge of the EcoCropsAID dataset lies in the substantial variability within the same category and the similarity between different categories. Moreover, the Google Earth program employs multiple remote imaging sensors, leading to variations in resolution quality, color, and contrast. In this dataset, the size of aerial images was standardized to 600x600 pixels with a resolution of 192 pixels per inch. The EcoCropsAID dataset contains a total of 5,400 images of economic crops.

- The EcoCropsAID dataset is not merely a collection of data but a catalyst for new research. It serves as an interdisciplinary resource across multiple domains, including remote sensing, geoinformatics, artificial intelligence, and computer vision. The dataset's unique features, such as the diverse phases captured in the aerial images and the challenge of information disparity, offer opportunities for researchers to propose novel approaches. These could involve extracting spatial and temporal features, developing new deep learning architectures, fusion models, or transformer-based approaches. By training on the EcoCropsAID dataset, researchers can achieve both high accuracy and efficient computational performance, opening new avenues for exploration and discovery.

- The EcoCropsAID dataset, freely available to the research community, has the potential to inspire researchers in multidisciplinary fields such as computer science and remote sensing. It provides a unique opportunity to develop novel classification algorithms for land use analysis, specifically focusing on economic crop aerial images, thereby contributing to advancements in these areas of research.

## BACKGROUND

Thailand's key economic crops, which are exported globally, include a variety of agricultural products such as rice, cassava, sugarcane, rubber, palm oil, longan, and animal-feed corn. Consequently, the cultivation of these crops has expanded across all regions of the country. This expansion is directly connected to land use planning, which is managed by the government and involves both public land management and the regulation of privately owned land [1,2]. The adoption of advanced technology to monitor crop cultivation provides an effective approach to optimizing land use and managing water resources, ensuring a reliable water supply for cultivated crops. In this research, the team focuses on exploring economic crop cultivation areas in the northeastern region of Thailand, as illustrated in Fig. 1, by utilizing deep learning algorithms to classify land use from satellite imagery, with a particular emphasis on five key crops: rice, sugarcane, cassava, rubber, and longan. To achieve this objective, data were collected using the Google Earth application, capturing aerial images obtained from various remote imaging sensors employed by Google for satellite imagery. Consequently, significant variations in resolution, color, and contrast are consistently observed in the aerial images.

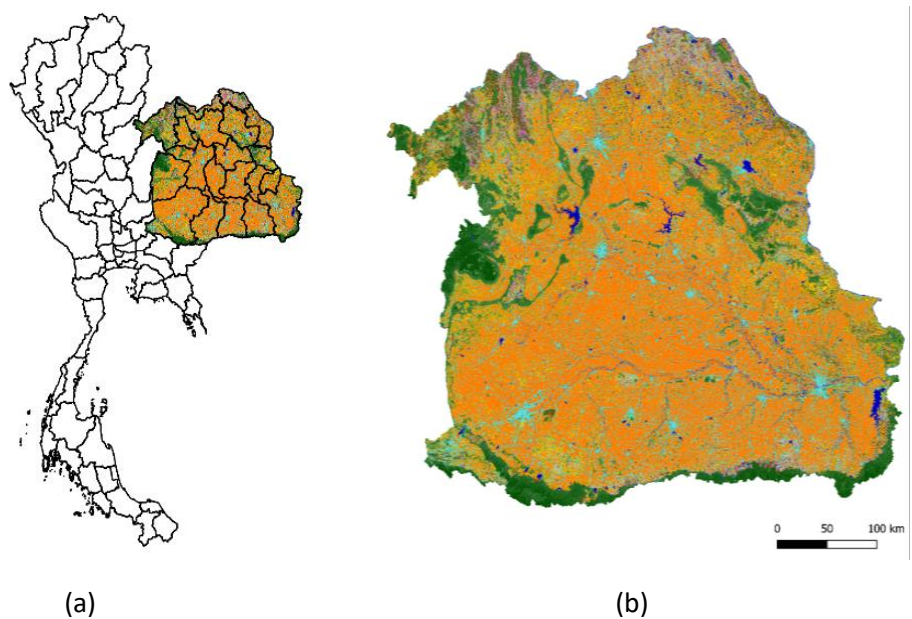

(a)                                        (b)

**Fig. 1.** Illustration of the northeastern region of Thailand: (a) the Thailand map and (b) the northeastern region. Note that the data. It is important to note that the map of the northeastern region of Thailand was generated using data provided by Potapov et al. [3].

## DATA DESCRIPTION

In Thailand, land use data can be accessed by contacting local government authorities; however, such information is typically restricted to specific areas. In contrast, comprehensive nationwide data is available through an online platform called Agri-Map Online, provided by the Ministry of Agriculture and Cooperatives. This platform allows users to easily access detailed land use information across the entire country. Agri-Map Online (https://agri-map-online.moac.go.th, see Fig. 2) serves as a crucial resource for researchers, policymakers, and agricultural planners by offering up-to-date data that supports informed decision-making and strategic planning. The platform's extensive coverage and accessibility make it an essential tool for analyzing land use trends, optimizing agricultural practices, and promoting sustainable development across the nation.

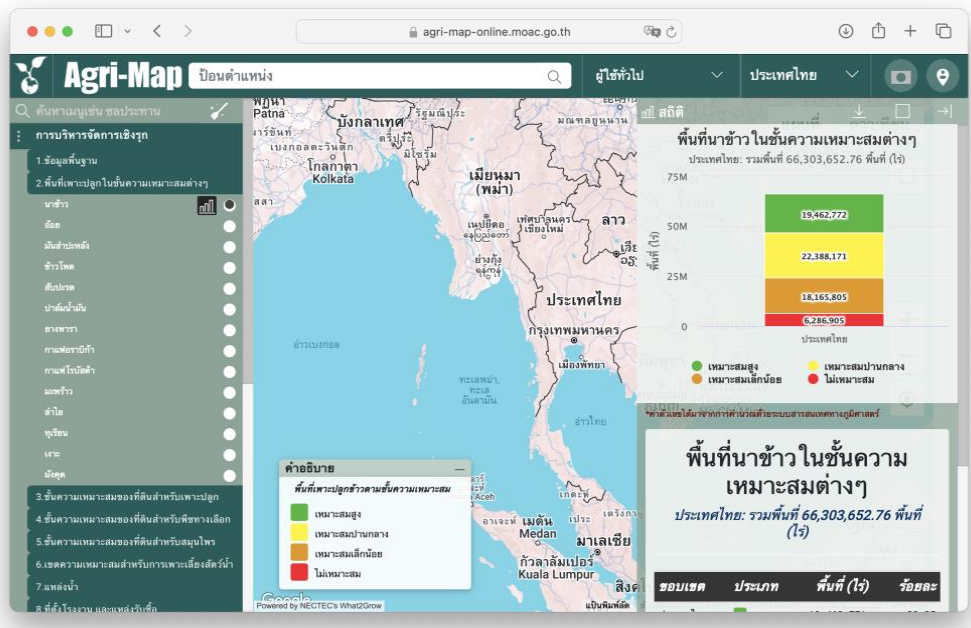

**Fig. 2.** Illustration of the Agri-Map Online platform, developed by the Ministry of Agriculture and Cooperatives in collaboration with the National Electronics and Computer Technology Center.

Following the validation of economic crop information using the Agri-Map Online platform, the research team selected satellite data from 2014 to 2018 using the Google Earth application, capturing various growth stages of each economic crop. Additionally, the Google Earth platform employs a range of remote imaging sensors, resulting in aerial images within the EcoCropAID dataset that exhibit significant variations in color, contrast, and resolution. Examples of aerial images captured through Google Earth are presented in Fig. 3.

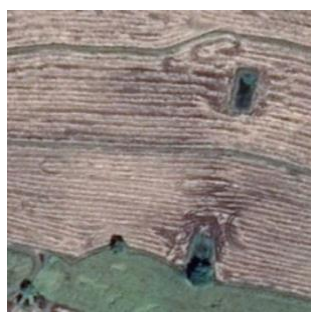  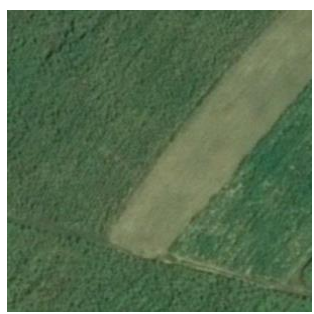  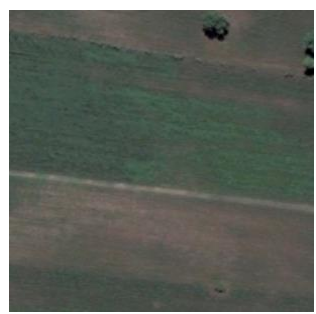

(a)                      (b)                      (c)

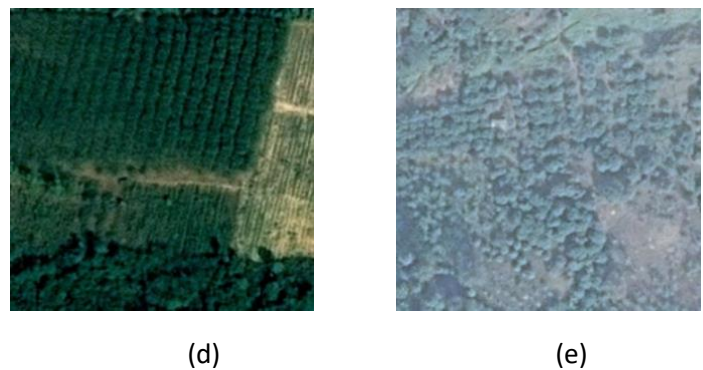

(d)                       (e)

**Fig. 3.** Illustration of aerial images of economic crops included in the EcoCropAID dataset: (a) rice, (b) sugarcane, (c) cassava, (d) rubber, and (e) longan.

The EcoCropAID dataset, collected in 2020, comprises 5,400 aerial images representing five key economic crops: rice, sugarcane, cassava, rubber, and longan [4]. These images display considerable variation in resolution, color, and contrast, posing a significant challenge for classification. The frequency distribution of aerial images across each economic crop is depicted in Fig. 4.

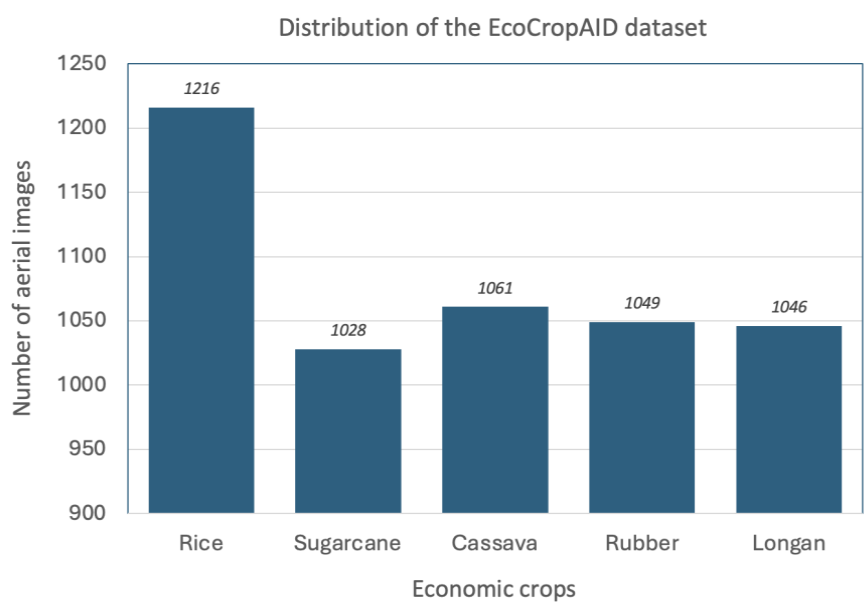

**Fig. 4.** Distribution of the EcoCropAID dataset.

Moreover, the dataset includes similarities between different categories (see Fig. 5(a) and Fig. 5(b)) and diverse patterns within the same category (see Fig. 5(c)), further complicating accurate classification.

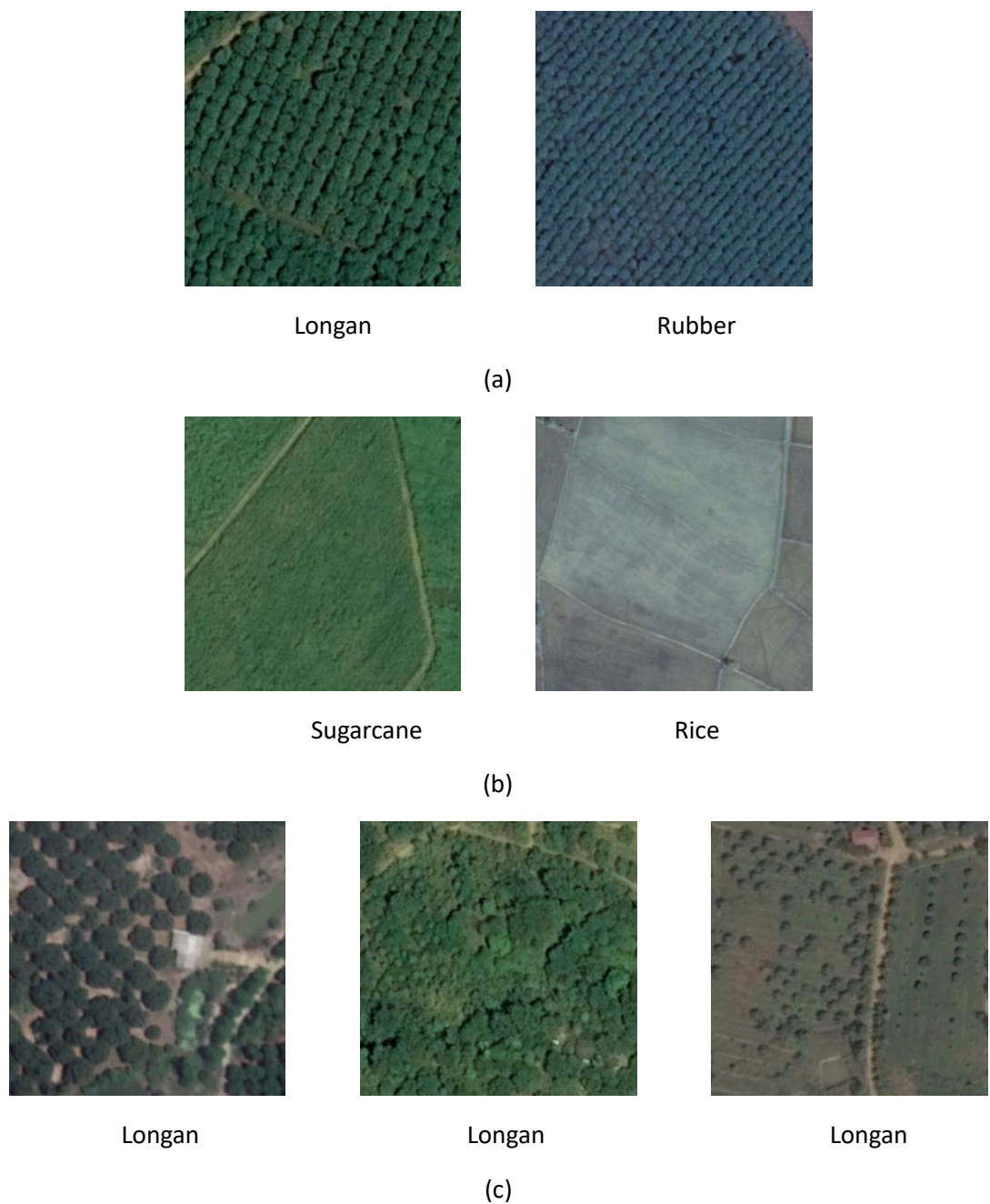

Longan                    Rubber

(a)

Sugarcane                    Rice

(b)

Longan                    Longan                    Longan

(c)

**Fig. 5.** Challenges in classifying aerial images in the EcoCropAID dataset: similarities between different categories, such as (a) longan and rubber and (b) sugarcane and rice, as well as diverse patterns within the same category, illustrated in (c) longan.

## EXPERIMENTAL DESIGN, MATERIALS AND METHODS

### Materials

The research team collected aerial images based on information provided by the Ministry of Agriculture and Cooperatives through the Agri-Map Online platform. The objective was to capture the growth stages of economic crops under various conditions. A total of 5,400 aerial images were randomly selected using the Google Earth application, considering a wide range of factors, including variations in resolution, color, and contrast. The EcoCropsAID dataset poses significant challenges for researchers developing algorithms to classify similar patterns across different categories and distinguish diverse patterns within the same category. The dataset serves as a valuable resource for advancing research in this domain and is publicly available through the Mendeley Data repository (https://data.mendeley.com/datasets/g8fhf7fbds/2) [5].

To collect the aerial images, the Google Earth application was utilized at a zoom level of 1:50 meters, with images captured at a resolution of 600×600 pixels, maintaining a 1:1 aspect ratio. The variations observed across each economic crop are depicted in Fig. 6 The images were then categorized into rice, sugarcane, cassava, rubber, and longan, with filenames starting with the crop category followed by a sequential number. The images were saved in JPG format with RGB channels. The file structure is presented in Fig. 7.

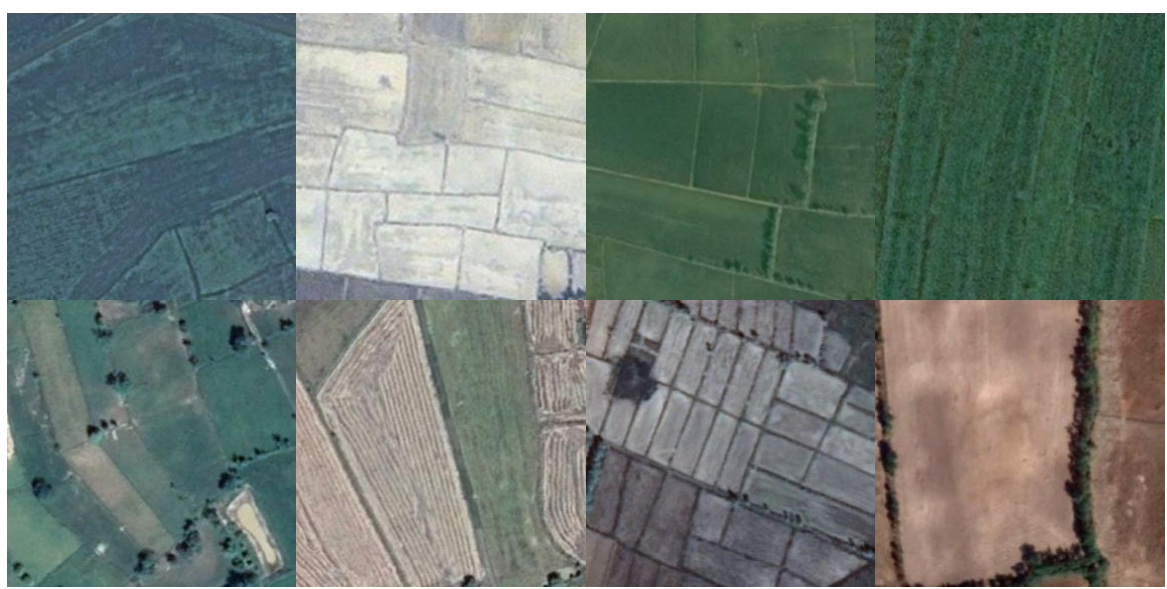

(a)

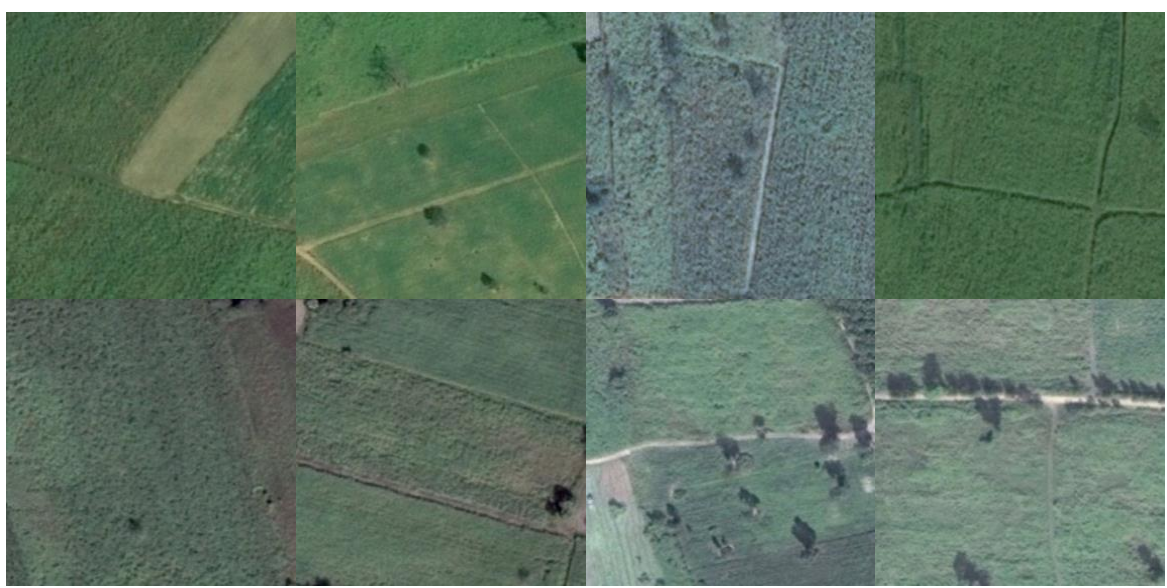

(b)

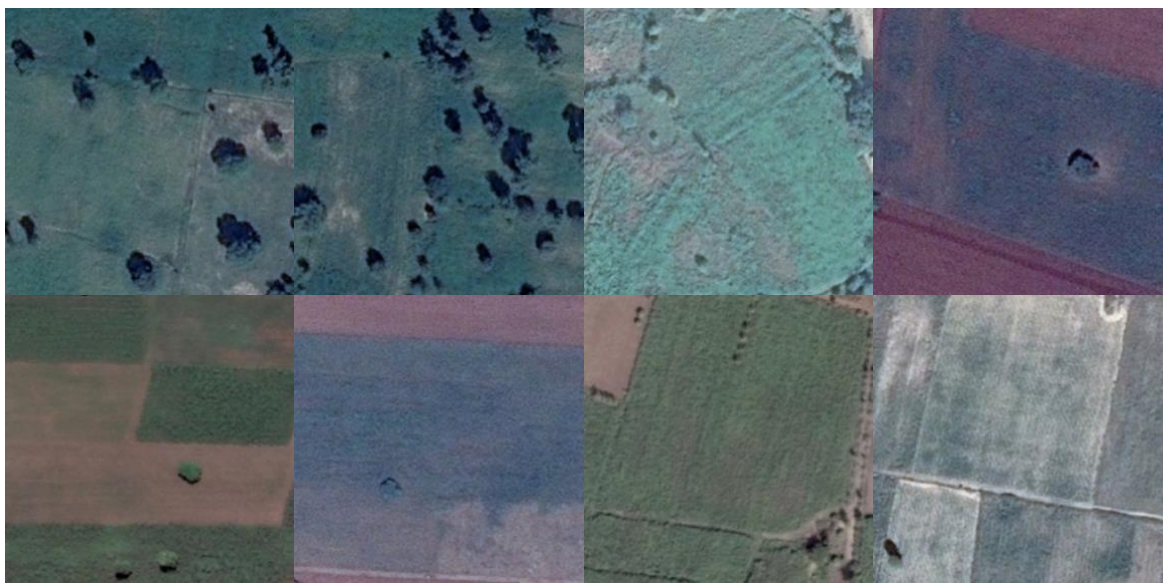

(c)

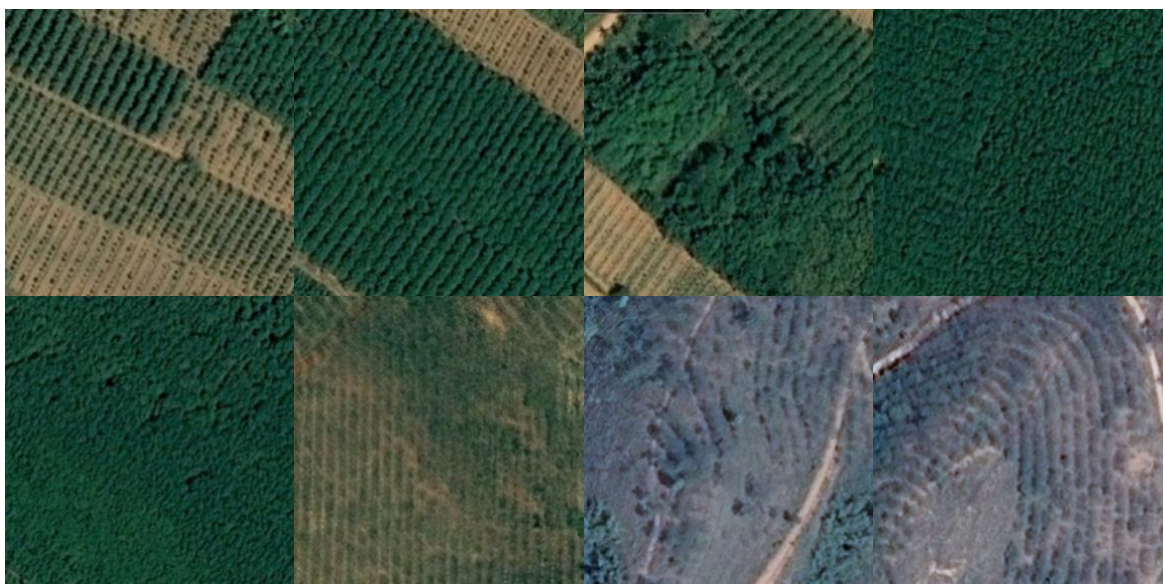

(d)

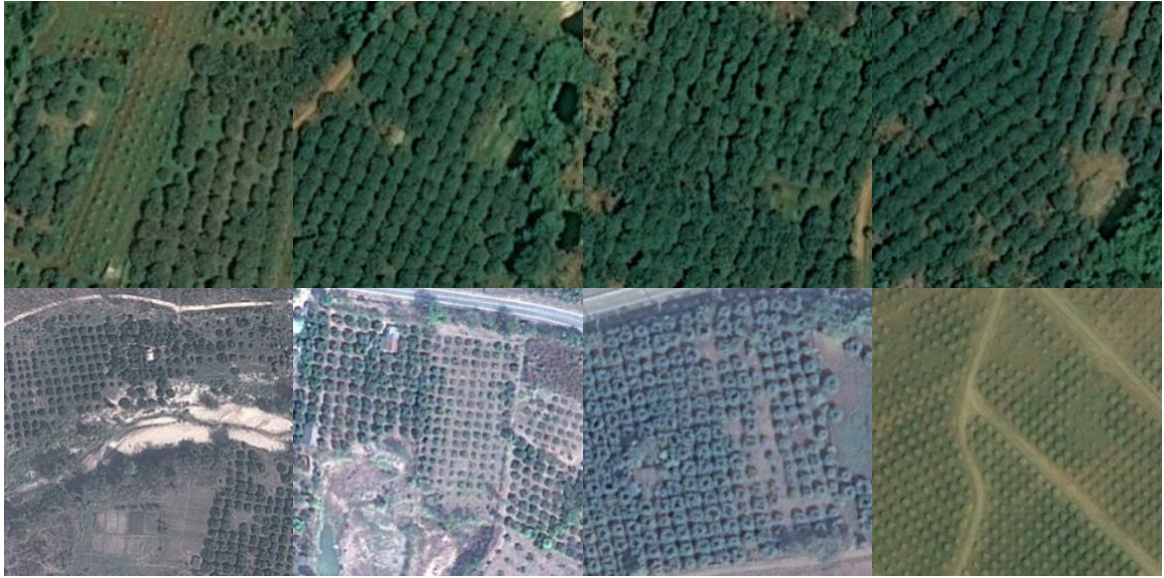

(e)

**Fig. 6.** Example of aerial images from the EcoCropAID dataset: (a) rice, (b) sugarcane, (c) cassava, (d) rubber, and (e) longan.

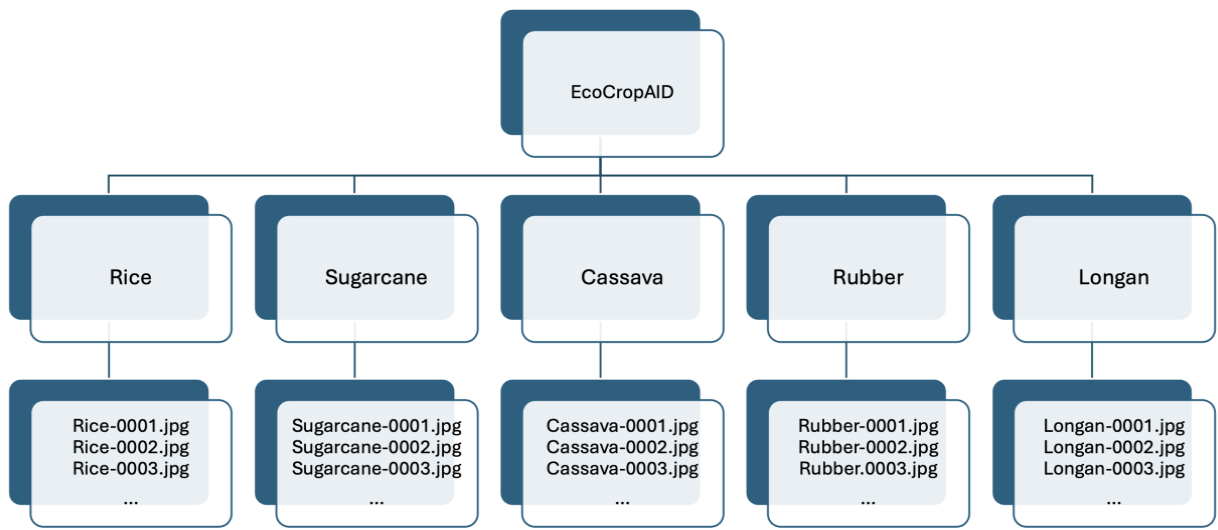

**Fig. 7.** Structure and format of files and filenames.

**Experimental design**

To develop the land use classification model for the AIWR dataset, Noppitak and Surinta [4] fine-tuned eight convolutional neural network models: VGG16, VGG19, Xception, ResNet152V2, InceptionResNetV2, MobileNetV2, DenseNet201, NASNetMobile, and NASNetLarge. Data augmentation techniques, including rotation and shift, were applied during model training. The three most optimal CNN models were then selected, and their output probabilities were combined using ensemble learning through the weighted average method. This approach resulted in an accuracy of 92.80% on the AIWR dataset. Although this accuracy is significant, it remains below the 95% threshold, indicating potential areas for further research where novel methods could be proposed to surpass this performance benchmark.

Classifying land use from aerial images poses significant challenges, primarily due to the diverse patterns within each category and the similarities across different classes. However, these challenges also present opportunities for advancing the field. Future land use classification approaches could focus on learning spatial and temporal features through multi-layer adaptive spatial-temporal feature fusion networks [6,7]. Additionally, vision transformer algorithms, which have proven effective in classifying remote sensing images [8,9], offer another promising solution that can enhance accuracy while optimizing computational efficiency.

## LIMITATIONS

Not applicable

## ETHICS STATEMENT

The authors have read and follow the ethical requirements and confirming that the current work does not involve human subjects, animal experiments, or any data collected from social media platforms.

## CRediT AUTHOR STATEMENT

**Sangdaow Noppitak:** Conceptualization, Data Curation, Investigation, Methodology, Resources, Validation, Writing – Original Draft; **Emmanuel Okafor:** Conceptualization, Validation, Writing – Review & Editing; **Olarik Surinta:** Supervision, Conceptualization, Experimental Design, Writing – Review & Editing, Funding Acquisition.

## ACKNOWLEDGEMENTS

This research project was financially supported by Mahasarakham University, Thailand.

## DECLARATION OF COMPETING INTERESTS

The authors declare that they have no known competing financial interests or personal relationships that could have appeared to influence the work reported in this paper.